\documentclass[10pt,twocolumn,letterpaper]{article}

\usepackage[pagenumbers]{cvpr} 

\definecolor{cvprblue}{rgb}{0.21,0.49,0.74}
\usepackage[pagebackref,breaklinks,colorlinks,allcolors=cvprblue]{hyperref}

\def\paperID{15634} 
\def\confName{CVPR}
\def\confYear{2026}

\title{RFAssigner: A Generic Label Assignment Strategy for Dense Object Detection}

\author{
\textbf{Ziqian Guan}$^{1,2}$\thanks{Equal contribution.}\quad
\textbf{Xieyi Fu}$^{2}$\footnotemark[1]\quad
\textbf{Yuting Wang}$^{1}$\quad
\textbf{Haowen Xiao}$^{1}$\quad
\textbf{Jiarui Zhu}$^{1}$\quad\\
\textbf{Yingying Zhu}$^{1}$\thanks{Corresponding author.}\quad
\textbf{Yongtao Liu}$^{2}$\footnotemark[2]\quad
\textbf{Lin Gu}$^{3}$\\[2pt]
\small $^{1}$ Guangzhou Institutes of Biomedicine and Health, Chinese Academy of Sciences\\
\small $^{2}$ North China Institute of Science and Technology\\
\small $^{3}$ RIKEN AIP, Japan\\[2pt]
}

\begin{document}
\maketitle

\begin{abstract}
Label assignment is a critical component in training dense object detectors. State-of-the-art methods typically assign each training sample a positive and a negative weight, optimizing the assignment scheme during training. However, these strategies often assign an insufficient number of positive samples to small objects, leading to a scale imbalance during training. To address this limitation, we introduce RFAssigner, a novel assignment strategy designed to enhance the multi-scale learning capabilities of dense detectors. RFAssigner first establishes an initial set of positive samples using a point-based prior. It then leverages a Gaussian Receptive Field (GRF) distance to measure the similarity between the GRFs of unassigned candidate locations and the ground-truth objects. Based on this metric, RFAssigner adaptively selects supplementary positive samples from the unassigned pool, promoting a more balanced learning process across object scales. Comprehensive experiments on three datasets with distinct object scale distributions validate the effectiveness and generalizability of our method. Notably, a single FCOS-ResNet-50 detector equipped with RFAssigner achieves state-of-the-art performance across all object scales, consistently outperforming existing strategies without requiring auxiliary modules or heuristics.
\end{abstract}
\section{Introduction}
\label{sec:intro}
Existing dense object detectors are predominantly categorized as either anchor-based or anchor-free. As elucidated by ATSS~\cite{atss}, the core distinction between these paradigms is their respective strategy for defining positive and negative training samples, a process that profoundly influences both training dynamics and final detection accuracy. Conventional anchor-based detectors typically rely on an Intersection-over-Union (IoU) criterion for assignment, where an anchor is matched to at most one ground-truth (GT) object, while a single GT may be assigned to multiple anchors. In contrast, anchor-free methods like AutoAssign~\cite{autoassign} employ dynamic, adaptive assignment strategies. AutoAssign, for instance, constrains positive sample centers to fall within GT boxes and assigns each sample both a positive and a negative weight, enabling the network to learn an optimal assignment end-to-end.

Although anchor-free approaches often surpass anchor-based methods on general-purpose benchmarks, their efficacy diminishes when detecting small objects. This performance degradation arises because small GT boxes have minimal spatial overlap with predefined anchors, and few, if any, feature map locations fall within their boundaries. While RFLA~\cite{rfla} has shown that using the matching degree between a Gaussian Receptive Field (GRF)~\cite{erf} and a GT object as an assignment metric can boost small-object detection, its underlying mechanism is problematic. RFLA effectively uses GRF priors to construct implicit anchor boxes, making it conceptually similar to traditional anchor-based methods. This design renders it incompatible with modern soft (non-binary) assignment strategies, thereby limiting its generalizability.

To overcome these challenges, we propose RFAssigner, a novel label assignment strategy founded on GRF principles. RFAssigner first initializes a set of candidate positive samples using point priors. It then refines this initial assignment using GRF priors, assigning each sample both a positive and negative weight to avoid hard binary decisions. This entire process is fully differentiable and can be optimized via backpropagation.

We conduct extensive validation of RFAssigner on three datasets with diverse object scale distributions: AI-TOD-v2~\cite{aitodv2}, MS-COCO-2017~\cite{coco}, and VisDrone-2019~\cite{visdrone2019}. We posit that RFAssigner is the first label assignment method explicitly designed for robust \textbf{cross-scale detection}. As our method is only active during the loss computation phase, it introduces no inference overhead. Evaluated under the standard 1$\times$ training schedule, RFAssigner consistently outperforms existing label assignment methods, demonstrating superior generalizability without recourse to auxiliary techniques.
\section{Related Work}
\label{related work}

\subsection{Object Detection}
The advent of deep learning has catalyzed a profound transformation in the field of object detection. Early methods, which relied on hand-crafted features within a sliding-window framework~\cite{histograms}, have been largely superseded. The introduction of Region-based Convolutional Neural Networks (R-CNN)~\cite{rcnn} marked a paradigm shift, leveraging CNNs for powerful feature extraction. Subsequent innovations, including Fast R-CNN~\cite{fastrcnn} and Faster R-CNN~\cite{fasterrcnn}, enhanced efficiency through shared feature computations and the integration of Region Proposal Networks (RPNs). Concurrently, single-stage detectors such as YOLO~\cite{yolo} and SSD~\cite{ssd} emerged, offering real-time performance by unifying localization and classification into a single pass. These seminal works have given rise to three dominant detector paradigms: anchor-based, anchor-free, and Transformer-based.

Anchor-based methods, pioneered by the RPN in Faster R-CNN~\cite{fasterrcnn}, utilize a predefined set of anchor boxes to guide object localization. This paradigm has been refined through techniques like Feature Pyramid Networks (FPNs)~\cite{fpn} for multi-scale feature fusion and focal loss~\cite{focalloss} for mitigating class imbalance. The YOLO series~\cite{yolo,yolov3,yolov4}, Cascade R-CNN~\cite{cascadercnn}, and ATSS~\cite{atss} have further advanced this line of research, though their performance can be sensitive to anchor design. In contrast, anchor-free methods eliminate the need for predefined anchors by predicting object properties directly. Notable examples include CornerNet~\cite{cornernet}, which detects pairs of object corners; CenterNet~\cite{centernet}, which identifies object centers; and FCOS~\cite{fcos}, which treats detection as a per-pixel prediction task. While simplifying the detection pipeline, these methods may face challenges in detecting small objects. More recently, Transformer-based architectures have enabled end-to-end detection. DETR~\cite{detr} introduced a set-prediction formulation using object queries, obviating the need for hand-crafted components like non-maximum suppression (NMS). Subsequent works such as Deformable DETR~\cite{detr} have improved computational efficiency, while Anchor DETR~\cite{anchordetr} and DAB-DETR~\cite{dabdetr} re-introduce learnable anchor priors. These models, however, typically require prolonged training schedules and operate under a sparse prediction paradigm.

\subsection{Label Assignment Strategies}
Label assignment, the process of designating training samples as positive or negative with respect to ground-truth (GT) objects, critically influences detector performance. Early strategies relied on static criteria, such as IoU thresholds~\cite{fasterrcnn} or spatial constraints~\cite{fcos}, which often result in suboptimal or imbalanced learning signals. To address this, dynamic assignment methods have been proposed. ATSS~\cite{atss} adaptively selects positive samples based on the statistical properties of IoU distributions, while PAA~\cite{paa} frames assignment as a probabilistic optimization problem. The concept of soft assignment further refines this process. GFL~\cite{gfocalv2} unifies classification scores with localization quality, and VFL~\cite{varifocalnet} introduces IoU-aware classification targets. Methods like AutoAssign~\cite{autoassign} and DW~\cite{dw} assign continuous weights instead of hard binary labels. However, these soft-assignment strategies may inadvertently allocate fewer positive samples to small objects, leading to scale-imbalanced learning. Although recent work like RFLA~\cite{rfla} has demonstrated that incorporating receptive-field information can benefit small-object detection, its design is not readily compatible with modern soft assignment frameworks.

\subsection{Challenges in Small Object Detection}
Detecting small objects presents a persistent challenge in computer vision, primarily due to the limited pixel information and consequently weak feature representations. A variety of techniques have been developed to mitigate this issue. Feature enhancement methods, such as FPN~\cite{fpn} and its variants~\cite{efficientdet}, aim to improve multi-scale feature fusion. Architecturally, models like S$^3$FD~\cite{s3fd} introduce specialized pyramid designs tailored for small-scale targets. Data augmentation strategies, including Mosaic~\cite{yolov4} and Copy-Paste~\cite{simple}, increase the frequency and diversity of small instances in the training data. Other research directions have explored the use of context modeling~\cite{li2021small}, attention mechanisms~\cite{wang2021small}, and specialized loss functions~\cite{kld}. Despite these advancements, detector performance on small objects remains constrained by a fundamental bottleneck: the scarcity of high-quality positive samples generated by conventional label assignment strategies. The development of dedicated benchmarks such as AI-TOD~\cite{aitodv2} and VisDrone~\cite{visdrone2019} continues to highlight the pressing need for methods explicitly engineered for this challenging scenario. Recent methods, such as RFLA\cite{rfla}, tailor their assignment strategies for small objects, while others like DQ-DETR\cite{dqdetr} incorporate specialized network modules. However, these specialized designs often significantly degrade the detector's performance on general-purpose benchmarks. Specifically, DQ-DETR\cite{dqdetr} requires extensive tuning of dataset-specific hyperparameters and incurs training times several multiples longer than standard schedules, hindering its generalizability.
\section{Method}
\begin{figure*}[t]
    \includegraphics[width=\linewidth]{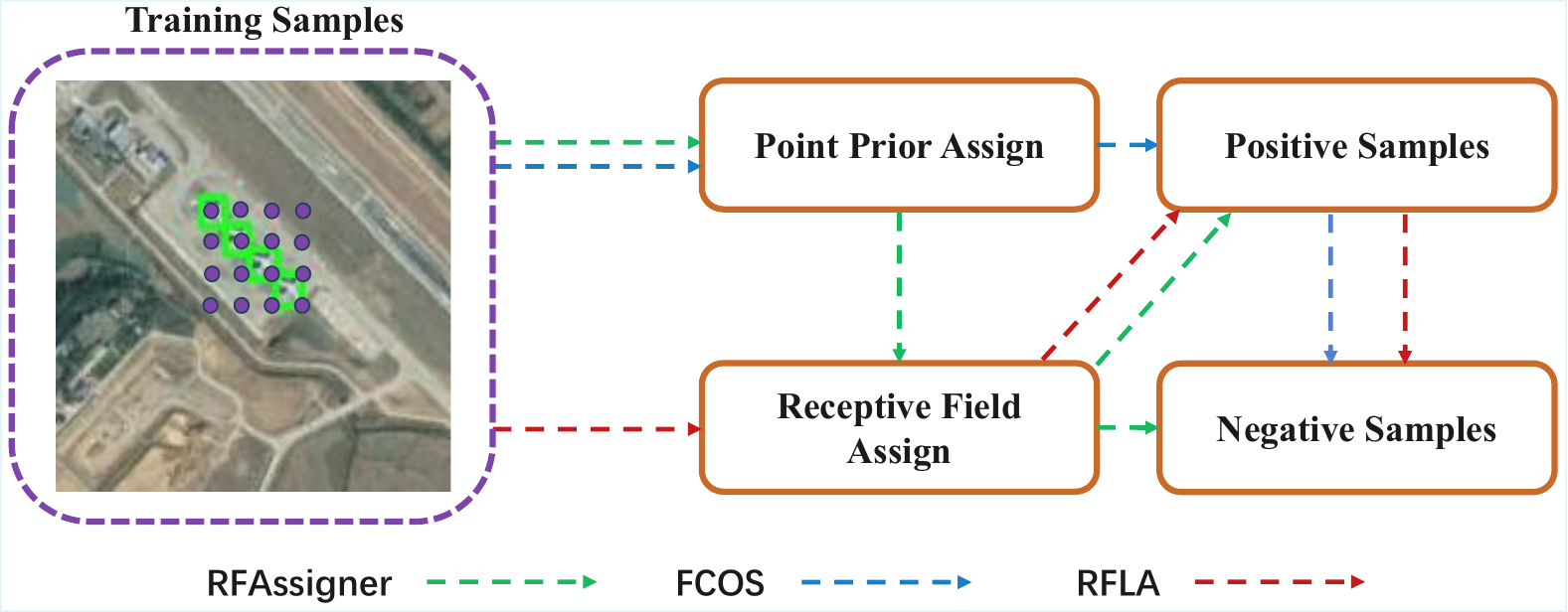}
    \caption{Comparison of label assignment pipelines. FCOS and RFLA define positive samples using a point prior and a GRF prior, respectively, with all other samples treated as negatives. RFAssigner synthesizes these approaches: it first initializes a positive set via a point prior (like FCOS), supplements this set using a GRF-based selection (inspired by RFLA), and finally assigns continuous positive and negative weights to all samples following the DW paradigm.}
    \label{fig:assigner_pipeline}
\end{figure*}

\subsection{Receptive Field Assignment}
Conventional label assignment frameworks define positive and negative samples primarily based on the spatial relationship between candidate locations and ground-truth (GT) boxes. This approach often leads to an insufficient allocation of positive samples for small objects, as the assignment rules are highly sensitive to minor spatial misalignments. RFLA~\cite{rfla} addresses this by using the correspondence between the Gaussian Receptive Fields (GRFs)~\cite{rfpnet} of feature map locations and GT boxes as the assignment criterion. However, this method fundamentally operates as an anchor-based hard-assignment strategy, wherein positive samples are explicitly defined and all other candidates are subsequently treated as negative. While this design enhances the detector's focus on small targets during training, it is incompatible with modern soft label assignment strategies like AutoAssign\cite{autoassign}, which degrades its performance on general-purpose object detection benchmarks.

In contrast, our method also models the receptive field (RF) of each feature point and the ground-truth (GT) box as a Gaussian distribution, but at the same time assigns an independent weight for positive and negative samples, thus avoiding hard assignment. Specifically, we utilize multiple RF scales to directly quantify their match with the GT and then dynamically select a subset of samples to augment an initial set of point-based priors. We model each GT bounding box as a 2D Gaussian distribution, where the mean corresponds to the box center and the covariance matrix encodes its extents:

\begin{equation}
    \label{gtmodel}
    \boldsymbol{\mu_{gt}} = \begin{bmatrix}
        x_{gt} \\
        y_{gt}
    \end{bmatrix},
    \quad
    \boldsymbol{\Sigma_{gt}} = \begin{bmatrix}
        \frac{w_{gt}^{2}}{4} & 0 \\ 
        0 & \frac{h_{gt}^{2}}{4}
    \end{bmatrix}.
\end{equation}

Analogously, the RF of each feature point is modeled as a 2D Gaussian distribution. The feature point's coordinates $(x_{\mathrm{tr}}, y_{\mathrm{tr}})$ serve as the mean vector. Departing from RFLA~\cite{rfla}, which uses half the Theoretical Receptive Field (TRF) radius, we define the diagonal entries of the covariance matrix using the full TRF radius:

\begin{equation}
    \label{trmodel}
    \boldsymbol{\mu}_{\mathrm{tr}} =
    \begin{bmatrix}
        x_{\mathrm{tr}} \\
        y_{\mathrm{tr}}
    \end{bmatrix},
    \quad
    \boldsymbol{\Sigma}_{\mathrm{tr}} =
    \begin{bmatrix}
        \frac{w_{\mathrm{tr}}^{2}}{4} & 0 \\
        0 & \frac{h_{\mathrm{tr}}^{2}}{4}
    \end{bmatrix}.
\end{equation}
Here, $w_{\mathrm{tr}}$ and $h_{\mathrm{tr}}$ denote the TRF diameters along the $x$- and $y$-axes.

RFLA~\cite{rfla} adopts the Kullback-Leibler Divergence (KLD)~\cite{kld} as its RF distance criterion (RFDC). Due to its scale-invariance, KLD is generally more suitable for handling objects of varying sizes than metrics like the Wasserstein Distance (WD)~\cite{wd}. However, KLD is asymmetric and can become unreliable when the two distributions have minimal overlap, potentially leading to suboptimal assignments. We therefore adopt the Gaussian Combined Distance (GCD)~\cite{gcd} as our RFDC to measure the correspondence between a feature point's RF and the GT. The GCD between the RF and GT Gaussian distributions is defined as:

\begin{equation}
    \begin{aligned}
        \mathbf{D}_{gc}^2\left(\mathcal{N}_{gt}, \mathcal{N}_{tr}\right) &= (\boldsymbol{\mu}_{gt} - \boldsymbol{\mu}_{tr})^{\top} 2 \boldsymbol{\Sigma}_{gt}^{-1} (\boldsymbol{\mu}_{gt} - \boldsymbol{\mu}_{tr}) \\
        &+ (\boldsymbol{\mu}_{tr} - \boldsymbol{\mu}_{gt})^{\top} 2 \boldsymbol{\Sigma}_{tr}^{-1} (\boldsymbol{\mu}_{tr} - \boldsymbol{\mu}_{gt}) \\
        &+ 2 (\boldsymbol{\Sigma}_{gt}^{-1/2})^{\top} \lVert \boldsymbol{\Sigma}_{gt}^{1/2} - \boldsymbol{\Sigma}_{tr}^{1/2} \rVert_{F}^{2} (\boldsymbol{\Sigma}_{gt}^{-1/2}) \\
        &+ 2 (\boldsymbol{\Sigma}_{tr}^{-1/2})^{\top} \lVert \boldsymbol{\Sigma}_{tr}^{1/2} - \boldsymbol{\Sigma}_{gt}^{1/2} \rVert_{F}^{2} (\boldsymbol{\Sigma}_{tr}^{-1/2}),
    \end{aligned}
    \label{eq:gcd}
\end{equation}
where $\left\| \cdot \right\|_{F}$ denotes the Frobenius norm.

GCD~\cite{gcd} is both scale-invariant and symmetric, and like WD, it provides a meaningful measure even for non-overlapping distributions. To normalize the distance into a similarity score, we apply an exponential transformation to map GCD to the range $(0,1)$, yielding our final receptive-field distance (RFD):

\begin{equation}
    \mathbf{RFD} = \exp\left( -\sqrt{\mathbf{D}_{gc}^{2}\left(\mathcal{N}_{gt}, \mathcal{N}_{tr}\right)} \right).
\end{equation}

\begin{figure*}[t]
    \includegraphics[width=\linewidth]{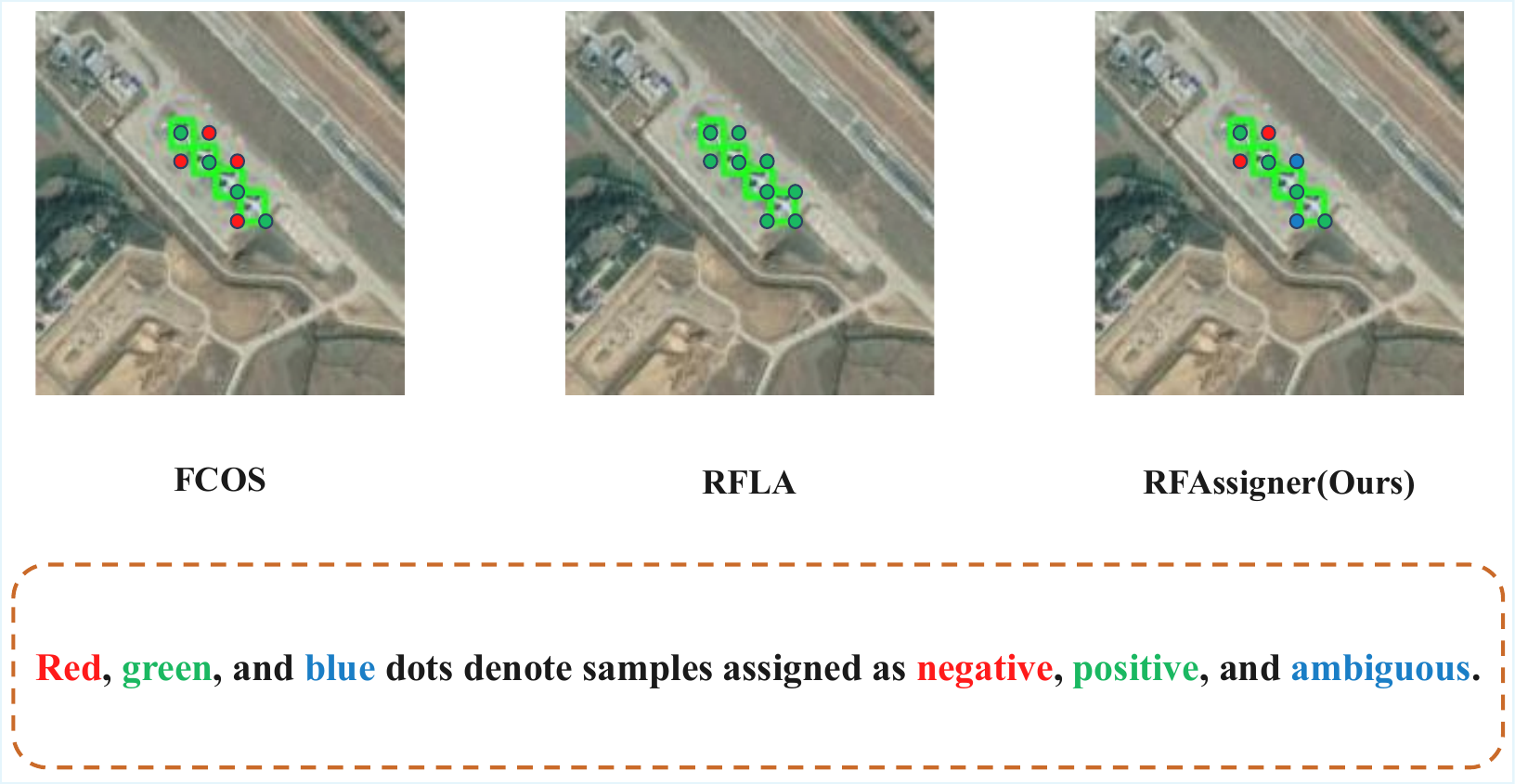}
    \caption{Visualization of different label assignment strategies. (Left) FCOS uses a point prior, assigning all locations within the GT box as positive (green). (Center) RFLA uses a hierarchical assignment based on RFD scores, which can designate locations outside the GT box as positive. (Right) RFAssigner begins with a point prior (green) and then adaptively selects ambiguous samples (blue) based on RFD statistics. These ambiguous samples are dynamically matched to GTs, allowing the assignment to be optimized throughout training.}
    \label{fig:assigner_cmp}
\end{figure*}

While RFLA~\cite{rfla} attempts to increase the number of positives for tiny targets via a hierarchical label assignment (HLA) based on ranked RFD scores, we find that HLA over-emphasizes these targets, which degrades performance on other scales and conflicts with soft assignment paradigms. AutoAssign~\cite{autoassign} introduced a point-prior paradigm with unique positive and negative weights per sample, and DW~\cite{dw} built upon this by decoupling the weight generation. However, both approaches can exacerbate scale imbalance. We argue that this scale imbalance originates from the monotonic process by which existing label assignment strategies define positive and negative sets.

To address this, we introduce \textbf{RFAssigner}, a novel receptive-field assignment strategy built upon the foundation of DW~\cite{dw}. As illustrated in Fig.~\ref{fig:assigner_pipeline}, RFAssigner first assigns an initial set of candidate positives using a point-prior rule. It then dynamically selects previously unassigned samples—based on the statistical properties of their RFD scores—to supplement this set. This supplementation primarily benefits smaller targets, as standard-sized objects typically receive sufficient candidates from the point-prior stage alone. Following RFLA~\cite{rfla}, we use four GRF scales ($1.0\times$, $0.75\times$, $0.50\times$, and $0.25\times$ the layer's TRF). To prevent the inclusion of low-quality samples, we select the top 9 candidates by RFD score, compute their mean $\mu$ and standard deviation $\sigma$, and add any candidate whose RFD exceeds $\mu + \sigma$ to the positive set.

\subsection{Ambiguous Matching}
To mitigate the negative effects of low-quality samples, RFLA~\cite{rfla} employs a hard threshold, labeling any sample whose maximum RFD to a GT is below 0.8 as background. This approach, however, disregards sample difficulty. Samples with high RFD scores (e.g., close to 1.0) are typically well-handled by point priors, and reassigning them can be counterproductive. Conversely, while candidates with very low RFDs are likely true negatives, this hard thresholding may neglect ambiguous samples that remain unassigned.

We therefore introduce an Ambiguous Matching strategy in RFAssigner, designed to target these difficult, unassigned samples. These ambiguous samples, which the detector cannot confidently classify, are dynamically assigned as positives to multiple GTs, thereby increasing the positive sample count, particularly for small objects.

Specifically, RFAssigner first generates a binary mask $M_p$ from the point-prior assignment. It then identifies ambiguous candidates by selecting samples whose RFD scores fall within a predefined range [0.60, 0.95]. Within this range, we rank the candidates by RFD and select the top-ranked ones to form a supplementary mask $M_{f}$. The final positive assignment mask $M_{result}$ is the union of the point-prior and supplementary masks, as formulated in Eq.~\ref{eq:match}. Crucially, Ambiguous Matching only modifies the positive branch (i.e., the samples used for positive loss and weight calculation); the negative branch continues to follow the center-prior mechanism of DW~\cite{dw}.

\begin{equation}
    M_{result} = M_{p} + M_{f} * (1 - M_{p})
    \label{eq:match}
\end{equation}

The total detection loss is composed of classification and regression terms as follows:
\begin{subequations}
\label{eq:total_loss}
\begin{align}
    \mathcal{L}_{det} &= \mathcal{L}_{cls} + \mathcal{L}_{reg}, \label{eq:loss_main} \\
    \mathcal{L}_{cls} &= \sum\nolimits_{n=1}^{N} \left[-w_{p o s}^{n} \ln \left(s^{n}\right)-w_{n e g}^{n} \ln \left(1-s^{n}\right)\right] \notag \\
    &\quad + \sum\nolimits_{m=1}^{M} F L\left(s^{m}, 0\right), \label{eq:loss_cls} \\
    \mathcal{L}_{reg} &= \sum\nolimits_{n=1}^{N} w_{pos}^{n}  \times GIoU\left(b, b^{\prime}\right), \label{eq:loss_reg}
\end{align}
\end{subequations}
where $N$ and $M$ are the numbers of positive and negative anchors defined by $M_{result}$, respectively, FL is the Focal Loss~\cite{focalloss}, GIoU is the regression loss~\cite{giou}, $s$ is the predicted \textit{cls} score, and $b$ and $b^{'}$ are the locations of the predicted box and the GT object, respectively.

\begin{figure*}[t]
	\centering
	\includegraphics[width=0.28\linewidth]{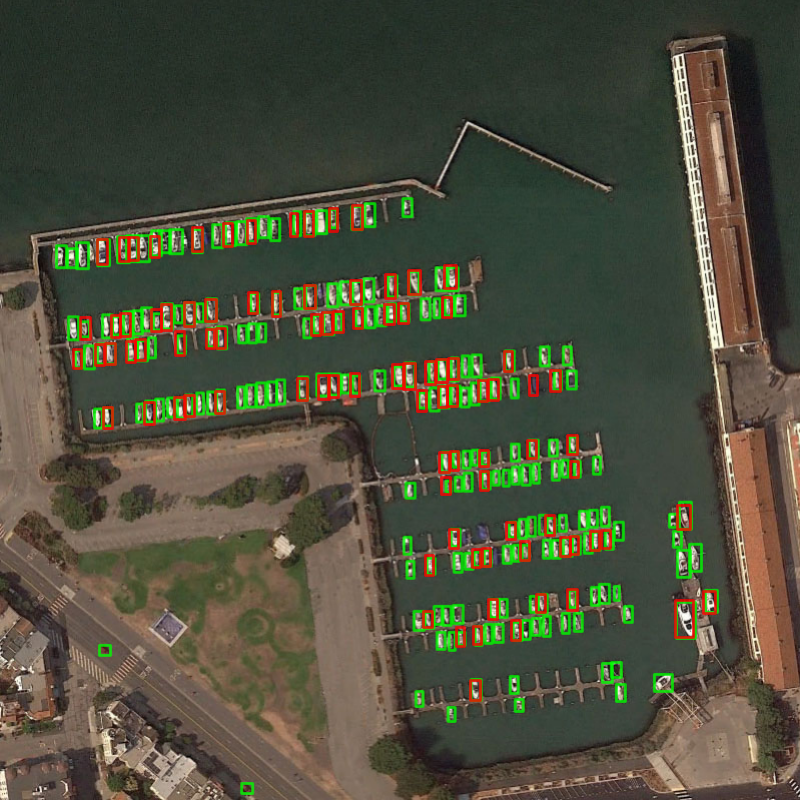}
	\includegraphics[width=0.28\linewidth]{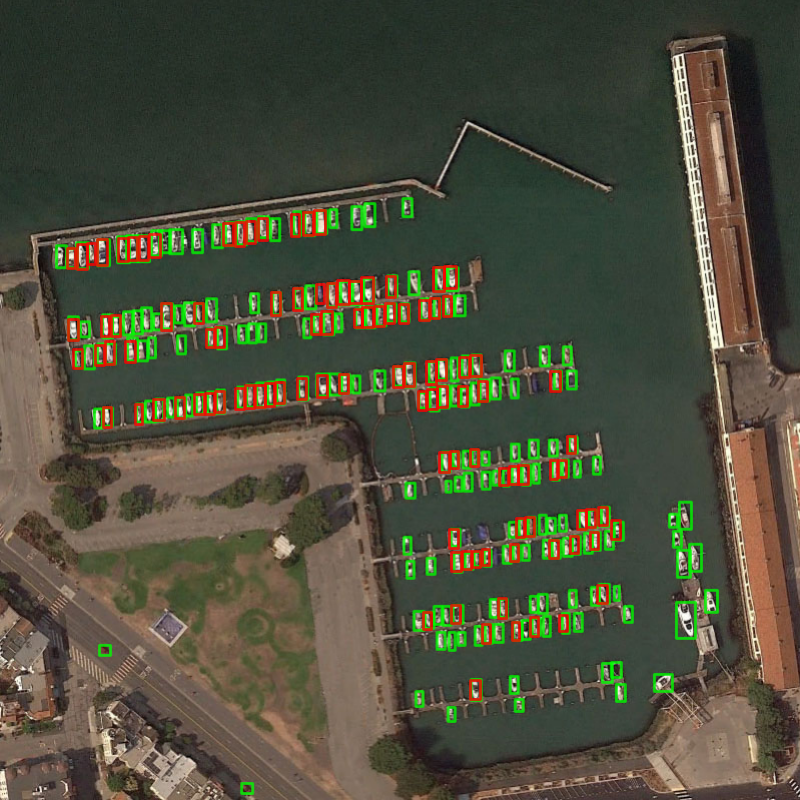}
	\includegraphics[width=0.28\linewidth]{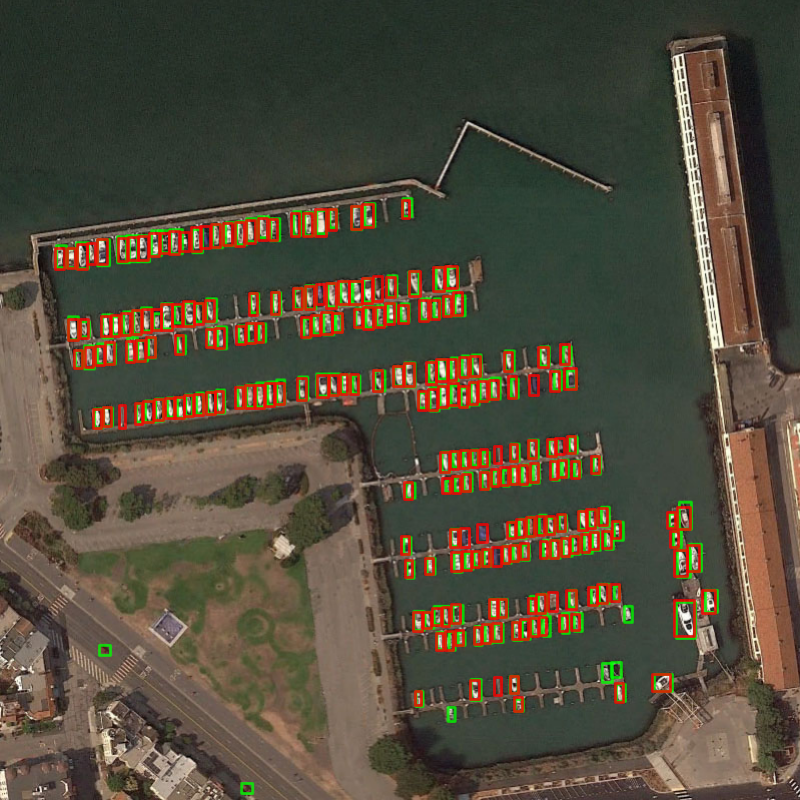}
	\includegraphics[width=0.28\linewidth]{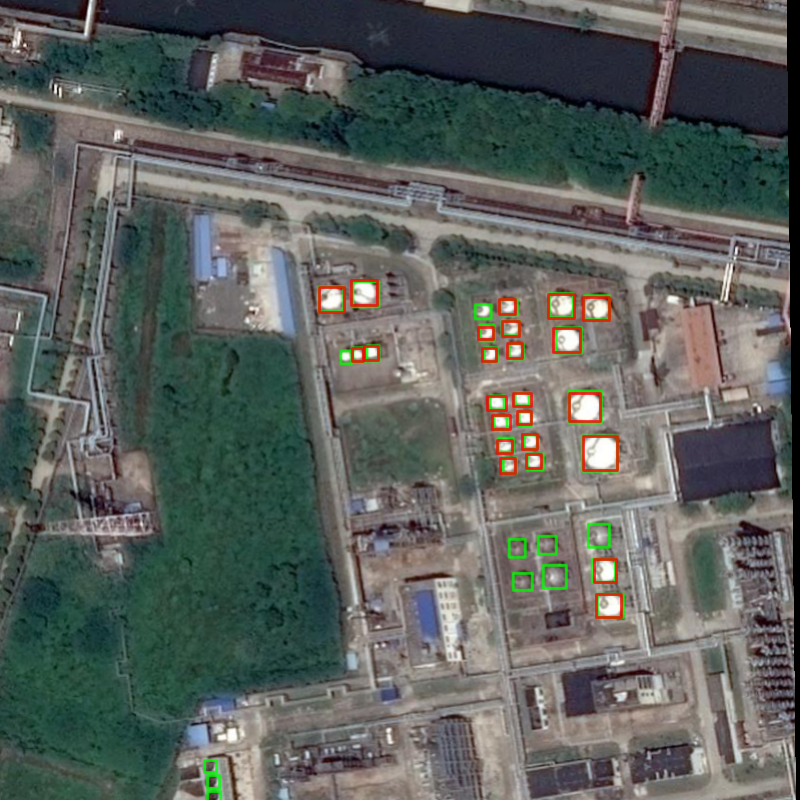}
	\includegraphics[width=0.28\linewidth]{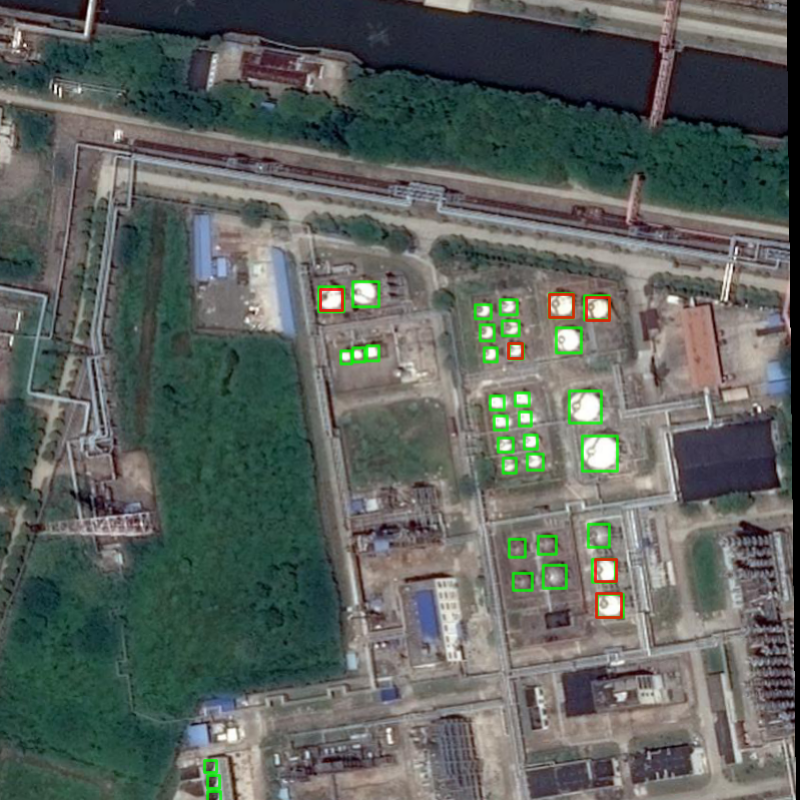}
	\includegraphics[width=0.28\linewidth]{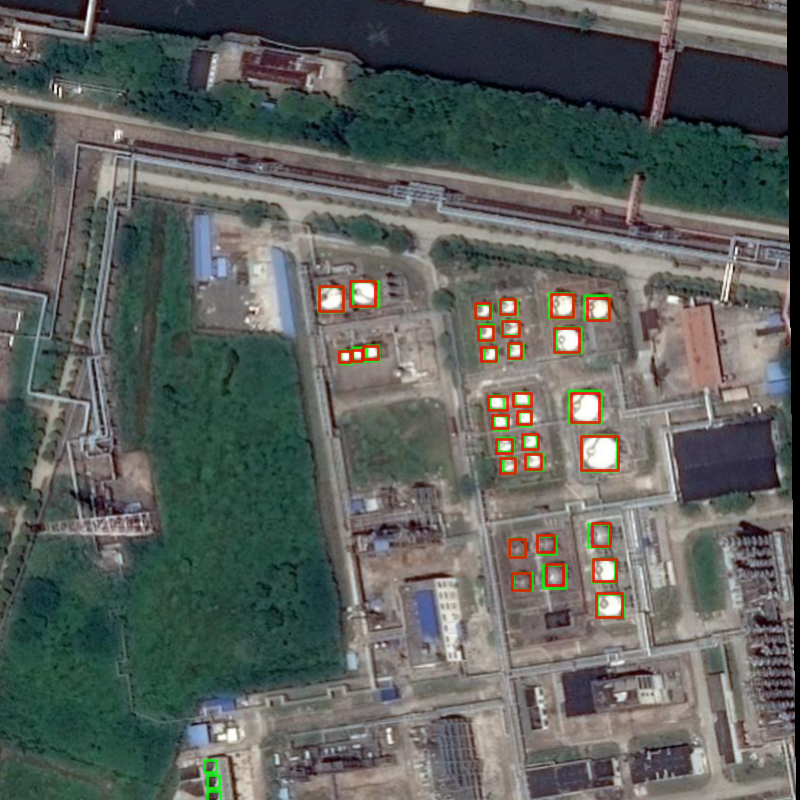}
	\caption{Qualitative detection results on the AI-TOD-v2 validation set. From left to right: DW, DW{*} + RFLA, and our RFAssigner{*}. Ground-truth boxes are shown in green, and predictions are in red. At a high confidence threshold (e.g., 0.5), DW{*} + RFLA does not consistently improve upon DW. In contrast, RFAssigner{*} demonstrates markedly superior detection performance.}
    \label{fig:visonreuslt}
\end{figure*}
\section{Experiments}
Our experimental evaluation is conducted on several benchmark datasets, including AI-TOD-v2\cite{aitodv2}, VisDrone-2019\cite{visdrone2019}, and MS-COCO-2017\cite{coco}. For the MS-COCO-2017 dataset, we employ the standard COCO Average Precision (AP) evaluation metrics. For the other datasets, we adhere to the evaluation protocols established for AI-TOD, which include multiple AP metrics: $\rm{AP}$, $\rm{AP_{0.5}}$, $\rm{AP_{vt}}$, $\rm{AP_{t}}$, $\rm{AP_{s}}$, and $\rm{AP_{m}}$. Specifically, $\rm{AP}$ denotes the mean Average Precision computed over IoU thresholds from 0.5 to 0.95 with a step of 0.05, and $\rm{AP_{0.5}}$ is the AP at a single IoU threshold of 0.5. Furthermore, $\rm{AP_{vt}}$, $\rm{AP_{t}}$, $\rm{AP_{s}}$, and $\rm{AP_{m}}$ represent the performance metrics for very tiny (2--8 pixels), tiny (8--16 pixels), and small objects, respectively. All experiments are implemented using the MMDetection\cite{mmdetection} framework. We consistently utilize a ResNet-50\cite{resnet} backbone, pre-trained on ImageNet\cite{imagenet} and augmented with a Feature Pyramid Network (FPN)\cite{fpn}. The models are trained for twelve epochs using an SGD\cite{sgd} optimizer with a momentum of 0.9, a weight decay of $10^{-4}$, and a batch size of 8 for all datasets. The initial learning rate is set to 0.01 and is decreased by a factor of ten at the eighth and eleventh epochs. During inference, bounding boxes with confidence scores below 0.05 are discarded, and Non-Maximum Suppression (NMS) is applied with an IoU threshold of 0.5.

\subsection{Datasets}
The Aerial Images Tiny Object Detection version 2 (AI-TOD-v2)\cite{aitodv2} dataset is a specialized benchmark for detecting minute objects in aerial imagery. It contains 700,621 object instances from eight categories across 28,036 images. The dataset is characterized by a mean object size of just 12.8 pixels, posing a significant challenge to detection algorithms. AI-TOD-v2 is a meticulously re-annotated version of its predecessor, designed to correct prevalent label noise and thereby enhance the training of tiny object detectors.

VisDrone-2019\cite{visdrone2019} comprises 261,908 video frames and 10,209 still images, capturing a wide diversity of scenes. The data spans 14 cities in China, features both urban and rural environments, includes various object types such as pedestrians and vehicles, and exhibits scene densities ranging from sparse to crowded.

The Microsoft Common Objects in Context (MS-COCO) 2017\cite{coco} dataset is a large-scale benchmark for object detection, segmentation, keypoint estimation, and image captioning. It contains approximately 330,000 images, where each image is annotated with 80 object categories and five descriptive captions, making it an invaluable resource for computer vision research.

\subsection{Ablation study}
All ablation studies are performed on the AI-TOD-v2\cite{aitodv2} dataset.

\textbf{Effectiveness of different RFD.} We evaluate the performance of RFAssigner using different metrics for the Ranking-based Feature Discrepancy (RFD). As reported in Table~\ref{tab:rfd}, Wasserstein Distance (WD) yields the weakest performance, which can be attributed to its lack of scale invariance. Although WD achieves the highest $\rm{AP}_{vt}$, its performance degrades for tiny ($\rm{AP}_{t}$) and larger object scales. In contrast, Kullback-Leibler Divergence (KLD), Normalized Wasserstein Distance (NWD), and Generalized Wasserstein Distance (GCD) are scale-invariant, which facilitates more effective learning of objects across various scales. GCD, which integrates the properties of both KLD and NWD, delivers the best overall performance. Therefore, we adopt GCD as the default RFD metric in all subsequent experiments.

\begin{table}[tb]
	\centering
	\caption{Comparison of Detection Performance with Varying different RFD.}
	\vspace{-2mm}
	\label{tab:rfd}
	\scalebox{0.85}{
	\setlength{\tabcolsep}{1.mm}
	\begin{tabular}{l|ccccccc}
	\toprule[1pt]
	Metrics & AP & $\rm{AP}_{0.5}$ & $\rm{AP}_{vt}$ & $\rm{AP}_{t}$ & $\rm{AP}_{s}$ & $\rm{AP}_{m}$ \\
	\hline
	WD\cite{wd}  & 16.4 & 40.7 & 5.3 & 15.7 & 21.9 & 27.7 \\
	KLD\cite{kld} & 17.3 & 43.1 & 4.3 & 17.4 & 22.0 & 30.0 \\
	NWD\cite{nwd} & 17.4 & 43.2 & 4.4 & 17.4 & 22.2 & 29.7 \\
	GCD\cite{gcd} & \textbf{17.8} & 44.2 & 5.2 & 17.4 & 22.8 & 29.9 \\
	\bottomrule[1pt]
	\end{tabular}
	}
\end{table}

\textbf{Ambiguous Matching Hyperparameters.} We conducted experiments to assess the robustness of the Ambiguous Matching module in RFAssigner to variations in its upper and lower threshold hyperparameters. The results in Table~\ref{tab:matching} show that excessively low thresholds cause an influx of low-quality samples during training, leading to a performance drop of approximately 1.0 AP. Conversely, as the upper threshold increases, the module's ability to focus on hard samples diminishes. Notably, setting the upper threshold to 1.0 and the lower to 0.6—effectively disabling ambiguous matching—results in a 1.0 AP performance loss. Based on these findings, we adopt an upper threshold of 0.95 and a lower threshold of 0.60 for subsequent experiments.

\begin{table}[tb]
	\centering
	\caption{Comparison of Detection Performance with Varying Upper and Lower Threshold Hyperparameters.}
	\vspace{-2mm}
	\label{tab:matching}
	\scalebox{0.8}{
	\setlength{\tabcolsep}{1.2mm}
	\begin{tabular}{c|cccc|cccc}
	\toprule[1pt]
	Upper Threshold & \multicolumn{4}{c|}{0.95} & 0.85 & 0.90 & 0.95 & 1.00 \\
	\hline
	Lower Threshold & 0.55 & 0.60 & 0.65 & 0.70 & \multicolumn{4}{c}{0.60} \\
	\hline
	AP & 16.7 & \textbf{17.8} & 17.3 & 16.5 & 16.3 & 17.2 & \textbf{17.8} & 16.8 \\ 
	$\rm{AP}_{0.5}$ & 42.7 & 44.2 & 44.1 & 41.3 & 41.3 & 42.7 & 44.2 & 42.7 \\
	$\rm{AP}_{vt}$ & 4.1 & 5.2 & 4.6 & 4.4 & 5.4 & 4.3 & 5.2 & 4.8 \\
	$\rm{AP}_{t}$ & 16.8 & 17.4 & 17.7 & 16.6 & 16.6 & 17.4 & 17.4 & 17.1 \\
	$\rm{AP}_{s}$ & 21.6 & 22.8 & 21.9 & 21.5 & 20.5 & 22.3 & 22.8 & 21.1 \\
	$\rm{AP}_{m}$ & 28.4 & 29.2 & 28.7 & 28.6 & 28.3 & 28.9 & 29.2 & 27.7 \\
	\bottomrule[1pt]
	\end{tabular}}
\end{table}
	
\subsection{Experiments on more datasets}

To validate the generalization capability of RFAssigner, we performed experiments on the VisDrone-2019 and MS-COCO-2017 datasets, with results presented in Tables~\ref{tab:vis2019} and~\ref{tab:coco2017}. The performance of DW\cite{dw} is known to degrade on small-scale objects. RFLA\cite{rfla}, which is primarily designed for tiny object detection (TOD), yields suboptimal results on standard-scale datasets and is incompatible with state-of-the-art soft label assignment strategies. In contrast, our proposed RFAssigner consistently achieves superior performance across datasets of varying object scales, demonstrating its strong generality. On MS-COCO-2017, RFAssigner improves upon DW by 0.9 AP in the $\rm{AP_{0.5}}$ metric, underscoring its robust capability for high-precision detection.

\begin{table}[tb]
	\centering
	\caption{Results of different LA strategies on VisDrone-2019 val set.}
	\vspace{-2mm}
	\label{tab:vis2019}
	\scalebox{0.85}{
	\setlength{\tabcolsep}{1.mm}
	\begin{tabular}{l|cccccc}
	\toprule[1pt]
	Method & AP & $\rm{AP}_{0.5}$ & $\rm{AP}_{vt}$ & $\rm{AP}_{t}$ & $\rm{AP}_{s}$ & $\rm{AP}_{m}$ \\
	\hline
	RetinaNet\cite{focalloss} & -- & 29.2 & -- & -- & -- & -- \\
	DCFL\cite{dcfl} & -- & 32.1 & -- & -- & -- & -- \\
	FCOS\cite{fcos} & 22.2 & 39.1 & 1.5 & 5.6 & 17.1 & 35.4 \\
	AutoAssign\cite{autoassign} & 25.0 & 46.0 & 2.7 & 9.5 & 20.4 & 37.6 \\
	DW\cite{dw} & 23.5 & 39.3 & 2.9 & 8.5 & 18.9 & 35.6 \\
	DW + RFLA\cite{rfla} & 25.4 & 46.0 & 3.8 & 10.4 & 21.8 & 37.7 \\
	RFAssigner(Ours) & \textbf{26.0} & 45.7 & 3.2 & 9.6 & 21.4 & 39.7 \\
	\bottomrule[1pt]
	\end{tabular}
	}
\end{table}

\begin{table}[tb]
	\centering
	\caption{Results of different LA strategies on MS-COCO-2017 val set.}
	\vspace{-2mm}
	\label{tab:coco2017}
	\scalebox{0.85}{
	\setlength{\tabcolsep}{1.mm}
	\begin{tabular}{l|cccccc}
	\toprule[1pt]
	Method & AP & $\rm{AP}_{0.5}$ & $\rm{AP}_{0.75}$ & $\rm{AP}_{s}$ & $\rm{AP}_{m}$ & $\rm{AP}_{l}$ \\
	\hline
	RetinaNet\cite{focalloss} & -- & 55.4 & -- & -- & -- & -- \\
	FCOS\cite{fcos} & 37.8 & 56.7 & 40.1 & 22.2 & 41.5 & 48.9 \\
	DCFL\cite{dcfl} & -- & 57.3 & -- & -- & -- & -- \\
	AutoAssign\cite{autoassign} & 40.2 & 59.7 & 43.2 & 22.9 & 43.7 & 52.6 \\
	DW\cite{dw} & 41.3 & 58.7 & 44.2 & 23.0 & 44.6 & 54.9 \\
	DW + RFLA\cite{rfla} & 37.4 & 56.4 & 39.9 & 21.9 & 41.2 & 47.6 \\
	RFAssigner(Ours) & \textbf{41.6} & 59.6 & \textbf{44.3} & 23.1 & 44.7 & 55.0 \\
	\bottomrule[1pt]
	\end{tabular}
	}
\end{table}
	
\subsection{Main results}

Table~\ref{tab:aitodv2} compares RFAssigner against state-of-the-art dense detectors and label assignment methods on the AI-TOD benchmark. Our RFAssigner* model achieves 22.3 AP, outperforming all competing single-stage detectors, with qualitative results shown in Figure~\ref{fig:visonreuslt}. Notably, even without leveraging the P2 feature level, the standard RFAssigner attains 17.8 AP, establishing a new state of the art among comparable single-stage methods.

\begin{table}[tb]
	\centering
	\caption{Results of different LA strategies on AI-TOD-v2 val set. Note that DW{*} + RFLA and RFAssigner{*} means using P2-P6 of FPN.}
	\vspace{-2mm}
	\label{tab:aitodv2}
	\scalebox{0.85}{
	\setlength{\tabcolsep}{1.mm}
	\begin{tabular}{l|cccccc}
	\toprule[1pt]
	Method & AP & $\rm{AP}_{0.5}$ & $\rm{AP}_{vt}$ & $\rm{AP}_{t}$ & $\rm{AP}_{s}$ & $\rm{AP}_{m}$ \\
	\hline
	RetinaNet\cite{focalloss} & 6.0 & 16.0 & 3.2 & 8.3 & 5.9 & 10.8 \\
	FCOS\cite{fcos} & 15.8 & 36.7 & 1.9 & 12.9 & 25.6 & 35.9 \\
	FCOS{*}\cite{fcos} & 17.1 & 40.1 & 5.4 & 17.4 & 22.8 & 27.2 \\
	DetectoRS\cite{detectors} & 12.9 & 27.7 & 0.1 & 8.0 & 26.3 & 41.0 \\
	ATSS\cite{atss} & 14.9 & 34.7 & 1.9 & 12.2 & 23.8 & 35.2 \\
	AutoAssign\cite{autoassign} & 16.7 & 44.3 & 4.0 & 16.3 & 22.1 & 28.5 \\
	DW\cite{dw} & 16.2 & 40.0 & 5.1 & 16.1 & 21.5 & 27.6 \\
	DW + RFLA\cite{rfla} & 16.8 & 42.9 & 4.9 & 16.4 & 23.2 & 26.9 \\
	RFAssigner(Ours) &  \textbf{17.8} & 44.2 & 5.2 & 17.4 & 22.8 & 29.9 \\
	DW{*} + RFLA & 21.1 & 50.9 & 6.9 & 21.5 & 26.2 & 34.0 \\
	RFAssigner{*}(Ours) & \textbf{22.3} & \textbf{53.0} & \textbf{7.5} & 22.2 & 27.1 & 35.6 \\
	\bottomrule[1pt]
	\end{tabular}
	}
\end{table}

\subsection{Discussion}
Further performance gains may be achievable by tuning the hyperparameters of RFAssigner to the specific characteristics of each dataset. Additional refinements to the label assignment architecture could also prove beneficial. Since the label assignment module operates exclusively during the training phase, RFAssigner introduces no additional computational overhead at inference. However, it does incur a minor increase in memory consumption during training, where the overhead is proportional to the number of anchor points.As current methods for receptive field calculation are tailored for standard convolutions, the applicability of RFAssigner is presently limited to Fully Convolutional Network (FCN)-based architectures.
\section{Conclusion}
In this work, we introduced RFAssigner, an adaptive label assignment paradigm for training precise, cross-scale dense object detectors. RFAssigner departs from conventional strategies by dynamically supplementing an initial set of positive samples, derived from point priors, with additional candidates selected based on Gaussian Receptive Field (GRF)~\cite{erf} priors. This allows for the dynamic assignment of distinct positive and negative weights to each training sample. Furthermore, we presented an Ambiguous Matching mechanism that directs the model's focus toward hard-to-classify samples while simultaneously filtering out low-quality candidates that could impede training. With a single FCOS-ResNet-50 detector, RFAssigner establishes a new state of the art, achieving 17.8 AP, 26.0 AP, and 41.6 AP on the AI-TOD-v2, VisDrone-2019, and MS-COCO-2017 datasets, respectively, without incurring any inference overhead. When leveraging finer-grained features from the P2--P6 levels of the FPN, our enhanced model, RFAssigner*, attains an impressive 22.3 AP on the challenging AI-TOD-v2 dataset. On the high-precision $\rm{AP}_{0.5}$ and tiny-object $\rm{AP}_{vt}$ metrics, RFAssigner{*} achieves performance on par with leading two-stage detectors. These results underscore the robust cross-scale detection capabilities of RFAssigner, a quality notably absent in prior label assignment methods.
{
    \small
    \bibliographystyle{ieeenat_fullname}
    \bibliography{main}
}


\end{document}